\documentclass[11pt,letterpaper]{article}
\usepackage[nohyperref]{emnlp2017}
\usepackage{times}
\usepackage{latexsym}
\usepackage{multirow}
\usepackage{amsfonts}
\usepackage{amssymb}
\usepackage{amsmath}
\usepackage{setspace}
\usepackage{enumitem}
\usepackage{tikz}
\usepackage{graphicx}

\usetikzlibrary{shapes}
\setlist{nosep} 

\DeclareMathOperator{\softmax}{softmax}

\emnlpfinalcopy

\def\emnlppaperid{399}

\title{Syllable-aware Neural Language Models:\\A Failure to Beat Character-aware Ones}

\author{Zhenisbek Assylbekov \\
        School of Science and Technology \\ Nazarbayev University\\ \tt{\normalsize zhassylbekov@nu.edu.kz}
        \And 
        Rustem Takhanov\\ School of Science and Technology \\ Nazarbayev University \\ \tt{\normalsize rustem.takhanov@nu.edu.kz}
        \AND
        Bagdat Myrzakhmetov \\ National Laboratory Astana \\ Nazarbayev University \\ \tt{\normalsize bagdat.myrzakhmetov@nu.edu.kz}
        \And
        Jonathan N. Washington \\ Linguistics Department\\ Swarthmore College \\ \tt{\normalsize jonathan.washington}\\ \tt{\normalsize@swarthmore.edu}}

\date{}

\begin{document}

\maketitle

\begin{abstract}
Syllabification does not seem to improve
word-level RNN language
modeling quality when compared to character-based segmentation. However, our best
syllable-aware language model, achieving performance comparable to the competitive character-aware
model, has 18\%--33\% fewer parameters and is
trained 1.2--2.2 times faster.
\end{abstract}

\section{Introduction}
Recent advances in neural language modeling  (NLM) are connected with character-aware models \cite{kim2016character,ling-EtAl:2015:EMNLP2,verwimp2017character}. This is a promising approach, and we propose the following direction related to it: We would like to make sure that in the pursuit of the most fine-grained representations one has not missed possible intermediate ways of segmentation, e.g., by syllables. Syllables, in our opinion, are better supported as linguistic units of language than single characters. In most languages, words can be naturally split into syllables:

\vspace{5pt}
ES: el par-la-men-to a-po-y\'o la en-mien-da\\
\indent RU: par-la-ment pod-der-\v{z}al po-prav-ku\\
\indent (EN: the parliament supported the amendment)
\vspace{5pt}

\noindent 
Based on this observation, we attempted to determine whether syllable-aware NLM has any advantages over character-aware NLM. We experimented with a variety of models but could not find any evidence to support this hypothesis: splitting words into syllables does not seem to improve the language modeling quality when compared to splitting into characters. However, there are some positive findings: while our best syllable-aware language model achieves performance comparable to the competitive character-aware model, it has 18\%--33\% fewer parameters and is 1.2--2.2 times faster to train. 

\section{Related Work}
Much research has been done on subword-level and subword-aware\footnote{\textit{Subword-level} LMs rely on subword-level inputs and make predictions at the level of subwords; \textit{subword-aware} LMs also rely on subword-level inputs but make predictions at the level of words.} neural language modeling  when subwords are characters \cite{ling-EtAl:2015:EMNLP2,kim2016character,verwimp2017character} or morphemes \cite{botha2014compositional,qiu2014co,cotterell2015morphological}. However, not much work has been done on syllable-level or syllable-aware NLM.  \newcite{mikolov2012subword} show that subword-level language models outperform character-level ones.\footnote{Not to be confused with character-aware ones, see the previous footnote.} They keep the most frequent words untouched and split all other words into syllable-like units. Our approach differs mainly in the following aspects: we make predictions at the word level, use a more linguistically sound syllabification algorithm, and consider a variety of more advanced  neural architectures.

We have recently come across a concurrent paper \cite{vania2017characters} where the authors systematically compare different subword units (characters, character trigrams, BPE \cite{DBLP:conf/acl/SennrichHB16a}, morphemes) and different representation models (CNN, Bi-LSTM, summation) on languages with various morphological typology. However, they do not consider syllables, and they experiment with relatively small models on small data sets (0.6M--1.4M tokens).

\section{Syllable-aware word embeddings}\label{syl_word_emb}
Let $\mathcal{W}$ and $\mathcal{S}$ be finite vocabularies of words and syllables respectively. We assume that both words and syllables have already been converted into indices. Let $\mathbf{E_\mathcal{S}}\in\mathbb{R}^{|\mathcal{S}|\times d_\mathcal{S}}$ be an embedding matrix for syllables --- i.e., it is a matrix in which the $s$th row (denoted as $\mathbf{s}$) corresponds to an embedding of the syllable $s\in \mathcal{S}$. Any word $w\in\mathcal{W}$ is a sequence of its syllables $(s_1, s_2, \ldots, s_{n_w})$, and hence can be represented as a sequence of the corresponding syllable vectors:
\begin{equation}
[\mathbf{s_1}, \mathbf{s_2}, \ldots, \mathbf{s_{n_w}}].\label{syl_seq}
\end{equation} 
The question is: How shall we pack the sequence (\ref{syl_seq}) into a single vector $\mathbf{x}\in\mathbb{R}^{d_\mathcal{W}}$ to produce a better embedding of the word $w$?\footnote{The same question applies to any model that segments words into a sequence of characters or other subword units.} In our case ``better'' means ``better than a character-aware embedding of $w$ via the Char-CNN model of \newcite{kim2016character}''. 
Below we present several viable approaches.

\begin{figure}
  
  \scalebox{.47}{
    \begin{tikzpicture}
      \node[text width=7cm] at (-0.1,3.5) {\huge unconstitutional}; 

      \node[draw,dotted] at (3.1,3.5) {\huge conditions};
      \node[text width=5cm] at (7.6,3.4) {\huge on};
      \draw[->] (3,2) -- (3,3);
      \draw[->] (-2,1) -- (-1,1);
      
      
      \draw [black] (-1,0) rectangle (1,2);
      [line width=20pt]
      \draw[->] (1,1) -- (2,1);
      \draw [black] (2,0) rectangle (4,2);
      \draw[->] (4,1) -- (5,1);
      \draw [black] (5,0) rectangle (7,2);
      \draw[->] (7,1) -- (8,1);
      \draw (9, 0) -- (9.5,0) -- (9.5,2) -- (9, 2);
      
      \node[text width=2.0cm, anchor=west, right] at (10,1)
    {stack of two LSTMs};
      
      \draw[->] (3,-0.8) -- (3,0);
      \draw[step=0.5 cm,black,very thin] (-1,-1) grid (7,-1.5);
      
      \draw[->] (3,-2.3) -- (3,-1.6);
      
      \draw (9, -1.5) -- (9.5,-1.5) -- (9.5,-1) -- (9, -1);
      \node[text width=7cm, anchor=west, right] at (10,-1.3)
    {word vector};
    
      \draw (-1, -3.0) -- (7,-3.0) -- (7,-2.5) -- (-1, -2.5) -- (-1, -3);
      \node[text width=7.5cm] at (5.0,-2.8)
    {\large Highway layers (optional)};
    
      \draw[->] (3,-3.8) -- (3,-3.1);
      \draw[step=0.5 cm,very thin] (-1,-4) grid (7,-4.5);
      
      
      \draw[->] (3,-5.3) -- (3,-4.6);
      \node [cloud, draw,cloud puffs=27,cloud puff arc=120, aspect=8.35, inner ysep=1em, ultra thick] at (3, -6.15) {};
      
      \node[text width=10cm] at (4.7,-6.2)
    {\large Syllable-aware word embedding model};
      
      \draw[->] (3,-7.8) -- (3,-7);
      
      \draw (1,-8) -- (1,-11.5);
      \draw[step=0.5 cm,black,very thin] (1.5,-8) grid (1,-11.5);
      
      \draw[step=0.5 cm,very thin] (1.5,-8) grid (1.5,-11.5);
      \draw[step= 0.5 cm,very thin] (2,-8) grid (2.5,-11.5);
      
      \draw[step= 0.5 cm,very thin] (2.5,-8) grid (2.5,-11.5);
      \draw[step=0.5 cm,very thin] (3,-8) grid (3.5,-11.5);
      
      \draw[step=0.5 cm,very thin] (3.5,-8) grid (3.5,-11.5);
      \draw[step=0.5 cm,very thin] (4,-8) grid (4.5,-11.5);
      
      \draw[step=0.5 cm,very thin] (5,-8) grid (5.5,-11.5);
      \draw[step=0.5 cm,very thin] (4.5,-8) grid (4.5,-11.5);
      
      \draw (9, -8) -- (9.5,-8) -- (9.5,-12) -- (9, -12);
      \node[text width=1.9cm, anchor=west, right] at (10,-10)
    {Syllable embeddings};
      
      \node[text width=2cm] at (2,-12) {\Large un};
      \node[text width=3cm] at (3.4,-12) {\Large con};
      \node[text width=3cm] at (4.47,-11.95) {\Large sti};
      \node[text width=3cm] at (5.5,-11.975) {\Large tu};
      \node[text width=3cm] at (6.3,-11.95) {\Large tional}; 
      
      \draw[->] (3,-13) -- (3,-12.3);
      
      \node[text width=7cm] at (1.25,-13.5) {\huge imposes};
      \node[draw,dotted] at (3,-13.4) {\huge unconstitutional};
      \node[text width=4cm] at (7.95,-13.4) {\huge conditions};
    \end{tikzpicture}
  }
  \caption{Syllable-aware language model.}
  \label{syl_lm}
\end{figure}
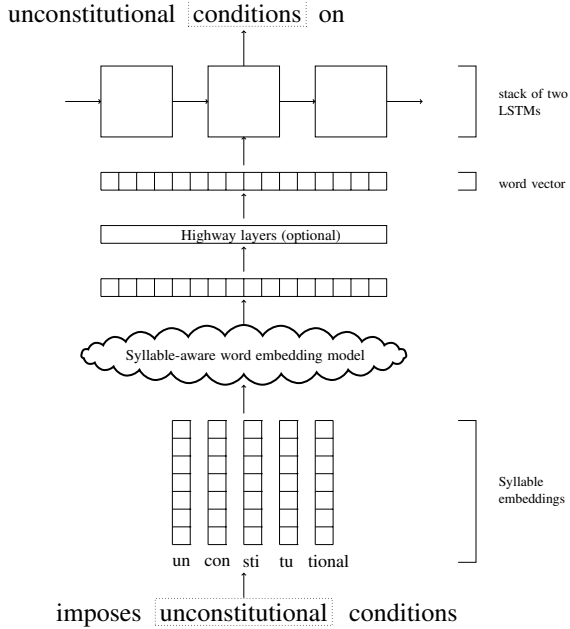

\subsection{Recurrent sequential model (Syl-LSTM)}\label{syl_lstm_dsc}
Since the syllables are coming in a sequence it is natural to try a recurrent sequential model:
\begin{equation}
\mathbf{h_t} = f(\mathbf{s_t}, \mathbf{h_{t-1}}),\qquad \mathbf{h_0} = \mathbf{0},\label{seq_model}
\end{equation}
which converts the sequence of syllable vectors (\ref{syl_seq}) into a sequence of state vectors $\mathbf{h_{1:n_w}}$. The last state vector $\mathbf{h_{n_w}}$ is assumed to contain the information on the whole sequence (\ref{syl_seq}), and is therefore used as a word embedding for $w$. There is a big variety of transformations from which one can choose $f$ in (\ref{seq_model}); however, a recent thorough evaluation \cite{jozefowicz2015empirical} shows that the LSTM \cite{hochreiter1997long} with its forget bias initialized to $1$ outperforms other popular architectures on almost all tasks, and we decided to use it for our experiments. We will refer to this model as \textit{Syl-LSTM}.

\subsection{Convolutional model (Syl-CNN)}
Inspired by recent work on character-aware neural language models \cite{kim2016character} we decided to try this approach (Char-CNN) on syllables. Our case differs mainly in the following two aspects:
\begin{enumerate}
\item The set of syllables $\mathcal{S}$ is usually bigger than the set of characters $\mathcal{C}$,\footnote{In languages with alphabetic writing systems.} and also the dimensionality $d_\mathcal{S}$ of syllable vectors is expected to be greater than the dimensionality $d_\mathcal{C}$ of character vectors. Both of these factors result in allocating more parameters on syllable embeddings compared to character embeddings.
\item On average a word contains fewer syllables than characters, and therefore we need narrower convolutional filters for syllables. This results in spending fewer parameters per convolution.
\end{enumerate}
This means that by varying $d_\mathcal{S}$ and the maximum width of convolutional filters $L$ we can still fit the parameter budget of \newcite{kim2016character} to allow fair comparison of the models.

Like in Char-CNN, our syllable-aware model, which is referred to as \textit{Syl-CNN-[L]}, utilizes max-pooling and highway layers \cite{srivastava2015training} to model interactions between the syllables. The dimensionality of a highway layer is denoted by $d_{\text{HW}}$.

\subsection{Linear combinations}
We also considered using linear combinations of syllable-vectors to represent the word embedding:
\begin{equation}\textstyle
\mathbf{x} = \sum_{t=1}^{n_w} \alpha_t(s_t) \cdot \mathbf{s_t}. \label{lin_comb}
\end{equation}
The choice for ${\alpha_t}$ is motivated mainly by the existing approaches (discussed below) which proved to be successful for other tasks.

\noindent\textbf{Syl-Sum:} Summing up syllable vectors to get a word vector can be obtained by setting $\alpha_t(s_t) = 1$. This approach was used by \newcite{botha2014compositional} to combine a word and its morpheme embeddings into a single word vector.

\noindent\textbf{Syl-Avg:} A simple average of syllable vectors can be obtained by setting $
\alpha_t(s_t) = {1}/{n_w}$. This can be also called a ``continuous bag of syllables'' in an analogy to a CBOW model \cite{mikolov2013efficient}, where vectors of neighboring words are averaged to get a word embedding of the current word.

\noindent\textbf{Syl-Avg-A:} We let the weights $\alpha_t$ in (\ref{lin_comb}) be a function of parameters $(a_1,\ldots,a_n)$ of the model, which are jointly trained together with other parameters. Here $n=\max_w\{n_w\}$ is a maximum word length in syllables. In order to have a weighted average in (\ref{lin_comb}) we apply a softmax normalization:
\begin{equation}
{\alpha_t} = \softmax(\mathbf{a})_t = \frac{\exp(a_t)}{\sum_{\tau=1}^n\exp(a_\tau)}\label{syl_avg_a}
\end{equation}

\noindent\textbf{Syl-Avg-B:} We can let  $\alpha_t$ depend on syllables and their positions:
$$
\alpha_t = \alpha_t(s_t)=\softmax(\mathbf{a}_{s_t} + \mathbf{b})_t
$$
where $\mathbf{A}\in\mathbb{R}^{d_\mathcal{S}\times n}$ (with elements ${a}_{s,t}$) is a set of parameters that determine the importance of each syllable type in each (relative) position,
$\mathbf{b}\in\mathbb{R}^{n}$ is a bias, which is conditioned only on the relative position. This approach is motivated by recent work on using an attention mechanism in the CBOW model \cite{ling2015not}.

We feed the resulting $\mathbf{x}$ from (\ref{lin_comb}) into a stack of highway layers to allow interactions between the syllables.

\subsection{Concatenation (Syl-Concat)}
In this model we simply concatenate syllable vectors (\ref{syl_seq}) into a single word vector:
$$\textstyle
\mathbf{x} = [\mathbf{s_1}; \mathbf{s_2}; \ldots; \mathbf{s_{n_w}}; \underbrace{\mathbf{0}; \mathbf{0}; \ldots; \mathbf{0}}_{n-n_w}]
$$
We zero-pad $\mathbf{x}$ so that all word vectors have the same length $n\cdot d_\mathcal{S}$ to allow batch processing, and then we feed $\mathbf{x}$ into a stack of highway layers.

\section{Word-level language model}
Once we have word embeddings $\mathbf{x_{1:k}}$ for a sequence of words $w_{1:k}$ we can use a word-level RNN language model to produce a sequence of states $\mathbf{h_{1:k}}$ and then predict the next word according to the probability distribution
$$
\Pr(w_{k+1}|w_{1:k})=\softmax(\mathbf{h_k}\mathbf{W} + \mathbf{b}),
$$
where $\mathbf{W}\in\mathbb{R}^{d_{\text{LM}}\times|\mathcal{W}|}$, $\mathbf{b}\in\mathbb{R}^{|\mathcal{W}|}$, and $d_{\text{LM}}$ is the hidden layer size of the RNN. Training the model involves minimizing the negative log-likelihood over the corpus $w_{1:K}$:
\begin{equation}\textstyle
-\sum_{k=1}^K\log\Pr(w_k|w_{1:k-1})\longrightarrow\min\label{nll}
\end{equation}
As was mentioned in Section \ref{syl_lstm_dsc} there is a huge variety of RNN architectures to choose from. The most advanced recurrent neural architectures, at the time of this writing, are 
recurrent highway networks \cite{zilly2016recurrent} and a novel model which was obtained through a neural architecture search with reinforcement learning \cite{zoph2016neural}. These  models can be spiced up with the most recent regularization techniques for RNNs \cite{gal2016theoretically} to reach state-of-the-art. 
However, to make our results directly comparable to those of \newcite{kim2016character} we select a two-layer LSTM and regularize it as in \newcite{zaremba2014recurrent}.

\section{Experimental Setup}\label{exp_setup}

We search for the best model in two steps: first, we block the word-level LSTM's architecture and pre-select the three best models under a small parameter budget (5M), and then we tune these three best models' hyperparameters under a larger budget (20M).

\noindent\textbf{Pre-selection:} We fix $d_\text{LM}$ (hidden layer size of the word-level LSTM) at 300 units per layer and run each syllable-aware word embedding method from Section \ref{syl_word_emb} on the English PTB data set \cite{marcus1993building}, keeping the total parameter budget at 5M. The architectural choices are specified in Appendix \ref{preselection}. 

\noindent\textbf{Hyperparameter tuning:} The hyperparameters of the three best-performing models from the pre-selection step are then thoroughly tuned on the same English PTB data through a random search according to the marginal distributions: 
\begin{itemize}
\item $d_\mathcal{S}\sim U(20, 650)$,\footnote{$U(a,b)$ stands for a uniform distribution over $(a,b)$.} 
\item $\log(d_\text{HW})\sim U(\log(160), \log(2000))$, 
\item $\log(d_\text{LM})\sim U(\log(300), \log(2000))$, 
\end{itemize}
with the restriction $d_\mathcal{S} < d_\text{LM}$. The total parameter budget is kept at 20M to allow for easy comparison to the results of \newcite{kim2016character}. Then these three best models (with their hyperparameters tuned on PTB) are trained and evaluated on small- (DATA-S) and medium-sized (DATA-L) data sets in six languages.

\noindent\textbf{Optimizaton} is performed in almost the same way as in the work of \newcite{zaremba2014recurrent}. See Appendix \ref{optimization} for details.

\noindent\textbf{Syllabification:} The true syllabification of a word requires its grapheme-to-phoneme conversion and then splitting it into syllables based on some rules. Since these are not always available for less-resourced languages, we decided to utilize Liang's widely-used hyphenation algorithm \cite{liang1983word}.

\section{Results}
The results of the pre-selection are reported in Table~\ref{pre_select}. 
\begin{table}[t]
\begin{center}
\begin{small}
\begin{tabular}{l c  l c}
\hline
Model & PPL & Model & PPL \\
\hline
LSTM-Word & 88.0 & Char-CNN  & 92.3 \\
\hline
Syl-LSTM  & 88.7 & Syl-Avg   & 88.5 \\
Syl-CNN-2 & 86.6 & Syl-Avg-A & 91.4\\
Syl-CNN-3 & \textbf{84.6} & Syl-Avg-B & 88.5 \\
Syl-CNN-4 & 86.8 & Syl-Concat & \textbf{83.7}\\
Syl-Sum   & \textbf{84.6} \\
\hline
\end{tabular}
\end{small}
\end{center}
\caption{Pre-selection results. PPL stands for test set perplexity, all models have $\approx5$M parameters.}
\label{pre_select}
\end{table}
All syllable-aware models comfortably outperform the Char-CNN when the budget is limited to 5M parameters. Surprisingly, a pure word-level model,\footnote{When words are directly embedded into $\mathbb{R}^{d_\mathcal{W}}$ through an embedding matrix $\mathbf{E_\mathcal{W}}\in\mathbb{R}^{|\mathcal{W}|\times d_\mathcal{W}}$.} LSTM-Word, also beats the character-aware one under such budget. The three best configurations are Syl-Concat, Syl-Sum, and Syl-CNN-3 (hereinafter referred to as Syl-CNN), and tuning their hyperparameters under 20M parameter budget gives the architectures in Table \ref{hyperparams-tuning}.
\begin{table}
\centering
\begin{small}
\begin{tabular}{l c c c c c c}
\hline
Model       & $d_\mathcal{S}$ & $d_\text{HW}$ & $d_\text{LM}$ & Size & PPL\\
\hline
Syl-CNN   & 242             & 1170          & 380           & 15M & 80.5\\
Syl-Sum     & 438             & 1256          & 435           & 18M & 80.3\\
Syl-Concat  & 228             & 781           & 439           & 13M & \textbf{79.4}\\
\hline
\end{tabular}
\caption{Hyperparameters tuning. In Syl-CNN, $d_\text{HW}$ is a function of the primary hyperparameter $c=195$ (see Appendix \ref{preselection}).}
\label{hyperparams-tuning}
\end{small}
\end{table}
The results of evaluating these three models on small (1M tokens) and medium-sized (17M--57M tokens) data sets against Char-CNN for different languages are provided in Table~\ref{evaluation}.
\begin{table}[t]
\setlength{\tabcolsep}{5pt}
\begin{center}
\begin{small}
\begin{tabular}{l c c c c c c c}
\hline
 Model   & EN & FR & ES & DE & CS & RU\\
\hline
 Char-CNN   & \textbf{78.9} & \textbf{184} & \textbf{165} & \textbf{239} & \textbf{371} & \textbf{261} & \parbox[t]{0.5mm}{\multirow{4}{*}{\rotatebox[origin=c]{90}{DATA-S}}}\\
Syl-CNN & 80.5 & 191 & 172 & 239 & 374 & 269 \\
Syl-Sum & 80.3 & 193 & 170 & 243 & 389 & 273 \\
Syl-Concat & 79.4 & 188 & 168 & 244 & 383 & 265 \\
\hline
Char-CNN& \textbf{160} & \textbf{124} & \textbf{118} & \textbf{198} & \textbf{392} & \textbf{190} & \parbox[t]{0.5mm}{\multirow{4}{*}{\rotatebox[origin=c]{90}{DATA-L}}}\\
Syl-CNN\footnotemark & -- & -- & -- & -- & -- & -- \\
Syl-Sum & 170 & 141 & 129 & 212 & 451 & 233 \\
Syl-Concat & 176 & 139 & 129 & 225 & 449 & 225 \\
\hline
\end{tabular}
\end{small}
\end{center}
\caption{Evaluation of the syllable-aware models against Char-CNN. In each case the smallest model, Syl-Concat, has 18\%--33\% less parameters than Char-CNN and is trained 1.2--2.2 times faster (Appendix \ref{size_speed}).}
\label{evaluation}
\end{table}
\footnotetext{Syl-CNN results on DATA-L are not reported since computational resources were insufficient to run these configurations.}
The models demonstrate similar performance on small data, but Char-CNN scales significantly better on medium-sized data. From the three syllable-aware models, Syl-Concat looks the most advantageous as it demonstrates stable results and has the least number of parameters. Therefore in what follows we will make a more detailed comparison of Syl-Concat with Char-CNN.

\begin{figure}[h]
\centering
\includegraphics[width=0.49\columnwidth]{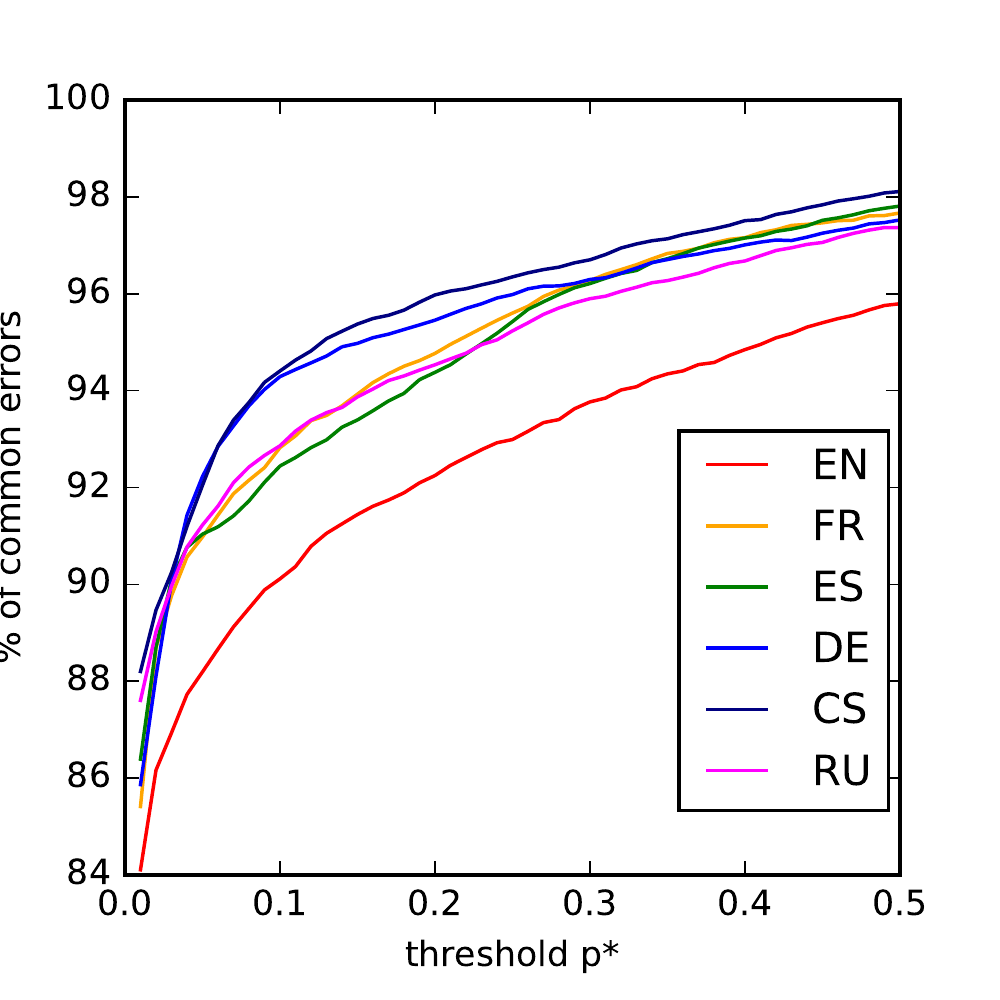}\includegraphics[width=0.49\columnwidth]{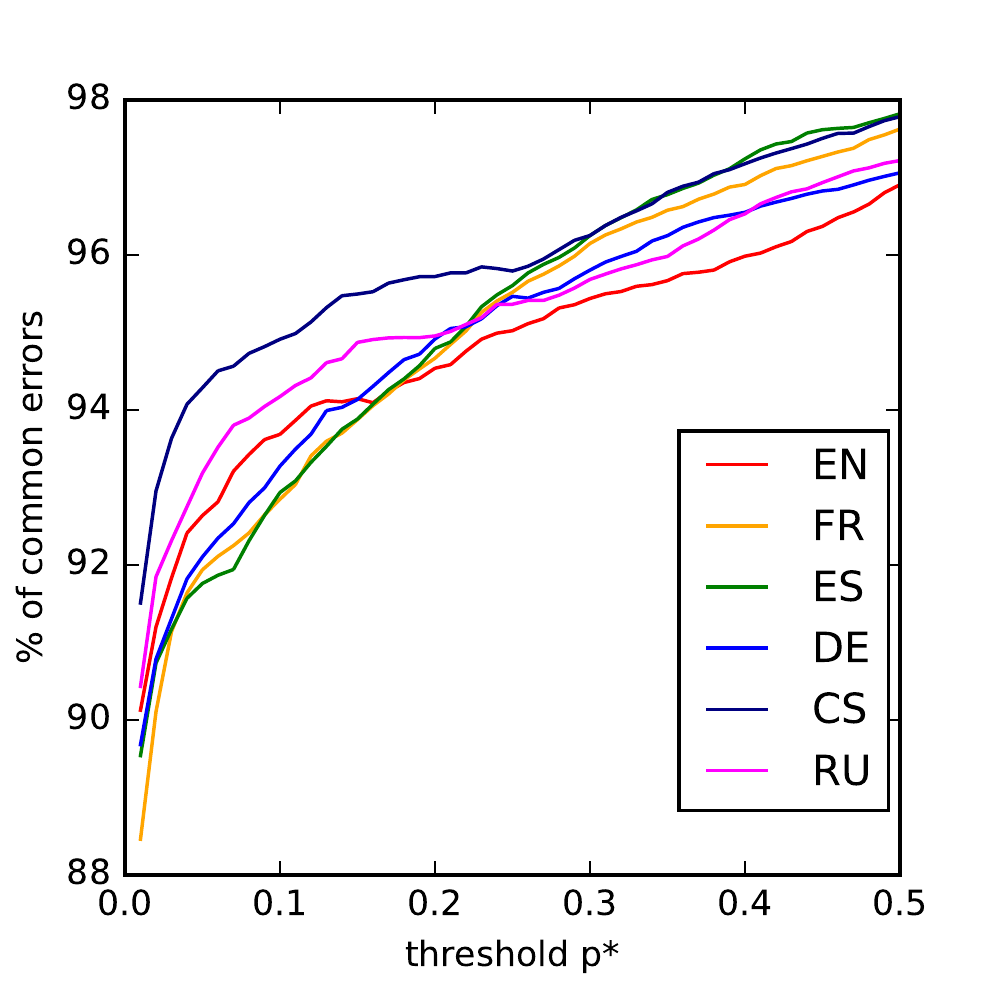}
\caption{Percentage of errors shared by both Syl-Concat and Char-CNN on DATA-S (left) and DATA-L (right).}
\label{shared_errors}
\end{figure}

\noindent\textbf{Shared errors:} It is interesting to see whether Char-CNN and Syl-Concat are making similar errors. We say that a model gives an error if it assigns a probability less than $p^\ast$ to a correct word from the test set. Figure \ref{shared_errors} shows the percentage of errors which are shared by Syl-Concat and Char-CNN depending on the value of $p^\ast$. 
We see that the vast majority of errors are shared by both models even when $p^\ast$ is small (0.01).

\begin{figure}[h]
\centering
\includegraphics[width=\columnwidth]{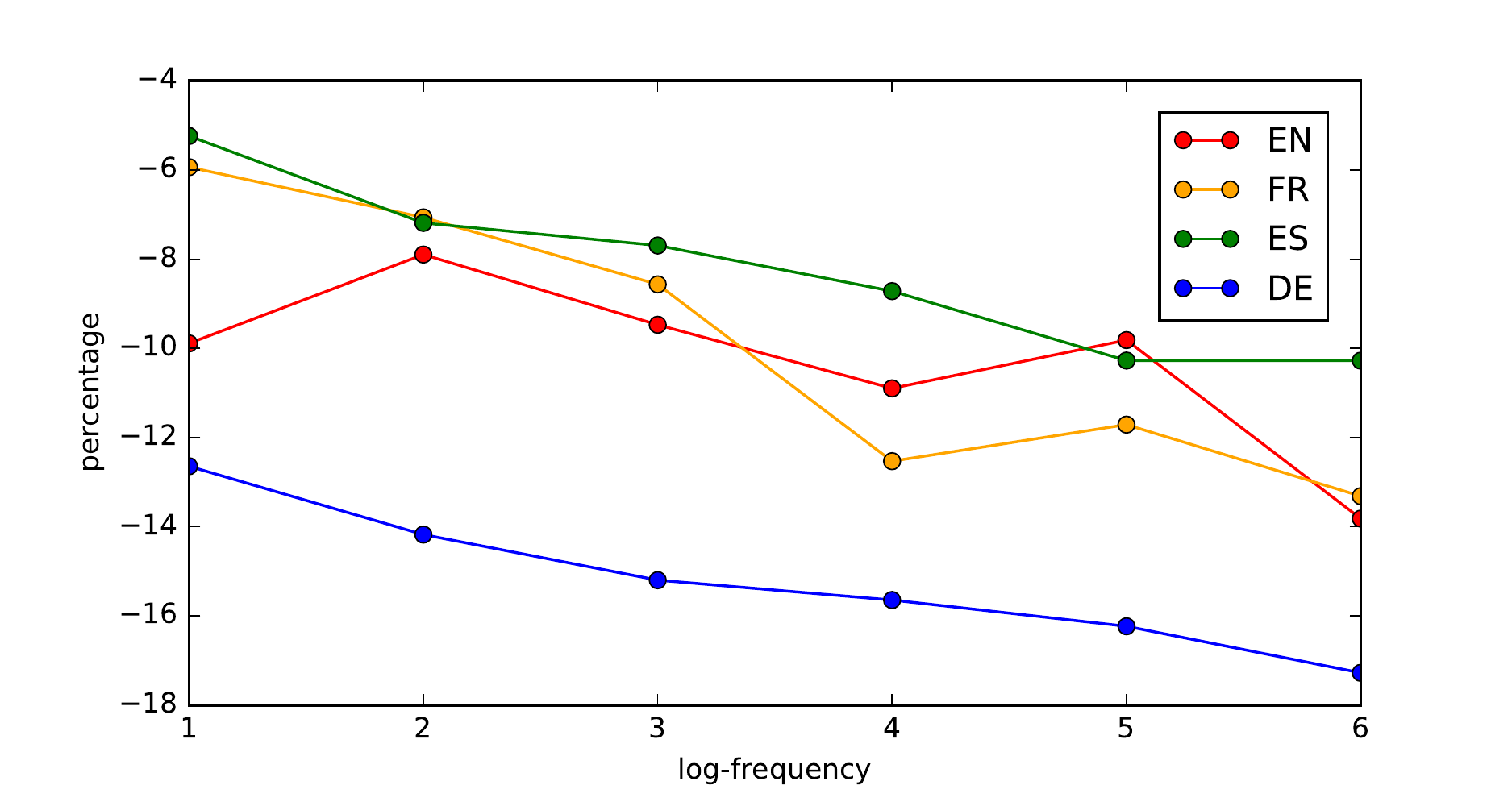}
\caption{PPL reduction by token frequency, Char-CNN relative to Syl-Concat on DATA-L.}
\label{ppl_reduct}
\end{figure}

\noindent\textbf{PPL breakdown by token frequency:} To find out \textit{how} Char-CNN outperforms Syl-Concat, we partition the test sets on token frequency, as computed on the training data. We can observe in Figure \ref{ppl_reduct} that, on average, the more frequent the word is, the bigger the advantage of Char-CNN over Syl-Concat. The more Char-CNN sees a word in different contexts, the more it can learn about this word (due to its powerful CNN filters). Syl-Concat, on the other hand, has limitations -- it cannot see below syllables, which prevents it from extracting the same amount of knowledge about the word. 

\noindent\textbf{PCA of word embeddings:} The intrinsic advantage of Char-CNN over Syl-Concat is also supported by the following experiment: We took word embeddings produced by both models on the English PTB, and applied PCA to them.\footnote{We equalized highway layer sizes $d_\text{HW}$ in both models to have same dimensions for embeddings. In both cases, word vectors were standardized using the z-score transformation.} Regardless of the threshold percentage of variance to retain, the embeddings from Char-CNN always have more principal components than the embeddings from Syl-Concat (see Table \ref{pca}).
\begin{table}
\begin{center}
\begin{tabular}{l c c c c}
\hline
Model      & 80\% & 90\% & 95\% & 99\% \\
\hline
Char-CNN   & 568  & 762  & 893  & 1038 \\
Syl-Concat & 515  & 729  & 875  & 1035 \\
\hline
\end{tabular}
\end{center}
\caption{Number of principle components when PCA is applied to word embeddings produced by each model, depending on \% of variance to retain.}
\label{pca}
\end{table}
This means that Char-CNN embeds words into higher dimensional space than Syl-Concat, and thus can better distinguish them in different contexts. 

\noindent\textbf{LSTM limitations:} During the hyperparameters tuning we noticed that increasing $d_\mathcal{S}$, $d_\text{HW}$ and $d_\text{LM}$ from the optimal values (in Table \ref{hyperparams-tuning}) did not result in better performance for Syl-Concat. Could it be due to the limitations of the word-level LSTM (the topmost layer in Fig.~\ref{syl_lm})? To find out whether this was the case we replaced the LSTM by a Variational RHN \cite{zilly2016recurrent}, and that resulted in a significant reduction of perplexities on PTB for both Char-CNN and Syl-Concat (Table \ref{rhn}). Moreover, increasing $d_\text{LM}$ from 439 to 650 did result in better performance for Syl-Concat. Optimization details are given in Appendix \ref{optimization}. 
\begin{table}[h]
\begin{center}
\begin{tabular}{l c c c c}
\hline
Model       & depth & $d_\text{LM}$ & Size & PPL \\
\hline
RHN-Char-CNN  & 8          & 650    & 20M & 67.6 \\
RHN-Syl-Concat& 8          & 439    & 13M & 72.0\\
RHN-Syl-Concat& 8          & 650    & 20M & 69.4\\
\hline
\end{tabular}
\caption{Replacing LSTM with Variational RHN.} 
\label{rhn}
\end{center}
\end{table}

\noindent\textbf{Comparing syllable and morpheme embeddings:} It is interesting to compare morphemes and syllables. We trained Morfessor 2.0 \cite{virpioja2013morfessor} in its default configuration on the PTB training data and used it instead of the syllabifier in our models. Interestingly, we got $\approx$3K unique morphemes, whereas the number of unique syllables was $\approx$6K. We then trained all our models on PTB under 5M parameter budget, keeping the state size of the word-level LSTM at 300 (as in our pre-selection step for syllable-aware models). The reduction in number of subword types allowed us to give them higher dimensionality $d_\mathcal{M} = 100$ (cf. $d_\mathcal{S} = 50$).\footnote{$\mathcal{M}$ stands for morphemes.}
 
Convolutional (Morph-CNN-3) and additive (Morph-Sum) models performed better than others with test set PPLs 83.0 and 83.9 respectively. Due to limited amount of time, we did not perform a thorough hyperparameter search under 20M budget. Instead, we ran two configurations for Morph-CNN-3 and two configurations for Morph-Sum with hyperparameters close to those, which were optimal for Syl-CNN-3 and Syl-Sum correspondingly. All told, our best morpheme-aware model is Morph-Sum with $d_\mathcal{M}=550$, $d_\text{HW}=1100$, $d_\text{LM}=550$, and test set PPL 79.5, which is practically the same as the result of our best syllable-aware model Syl-Concat (79.4). 
This makes Morph-Sum a notable alternative to Char-CNN and Syl-Concat, and we defer its thorough study to future work.

\noindent\textbf{Source code:} The source code for the models discussed in this paper is available at  https://github.com/zh3nis/lstm-syl.

\section{Conclusion}
It seems that syllable-aware language models fail to outperform competitive character-aware ones. However, usage of syllabification can reduce the total number of parameters and increase the training speed, albeit at the expense of language-dependent preprocessing. Morphological segmentation is a noteworthy alternative to syllabification: a simple morpheme-aware model which sums morpheme embeddings looks promising, and its study is deferred to  future work.

\appendix

\section{Pre-selection}\label{preselection}
In all models with highway layers there are two of them and the non-linear activation of any highway layer is a ReLU.

\noindent\textbf{LSTM-Word:} $d_\mathcal{W}=108$, $d_{\text{LM}}=300$.

\noindent\textbf{Syl-LSTM:} $d_\mathcal{S}=50$, $d_{\text{LM}}=300$.

\noindent\textbf{Syl-CNN-[$L$]:} $d_\mathcal{S}=50$, convolutional filter widths are $[1,\ldots, L]$, the corresponding convolutional filter depths are $[c\cdot l]_{l=1}^L$, $d_{\text{HW}} = c\cdot(1+\ldots+L)$. We experimented with $L=2,3,4$. The corresponding values of $c$ are chosen to be $120, 60, 35$ to fit the total parameter budget. CNN activation is $\tanh$.

\noindent\textbf{Linear combinations:} We give higher dimensionality to syllable vectors here (compared to other models) since the resulting word vector will have the same size as syllable vectors (see (\ref{lin_comb})). $d_\mathcal{S}=175$, $d_\text{HW}=175$ in all models except the Syl-Avg-B, where we have $d_\mathcal{S}=160$, $d_\text{HW}=160$. 

\noindent\textbf{Syl-Concat:} $d_\mathcal{S}=50$, $d_\text{HW}=300$.

\section{Optimization}\label{optimization}
\textbf{LSTM-based models:} We perform the training (\ref{nll}) by truncated BPTT \cite{werbos1990backpropagation,graves2013generating}. We backpropagate for 70 time steps on DATA-S and for 35 time steps on DATA-L using stochastic gradient descent where the learning rate is initially set to 1.0 and halved if the perplexity does not decrease on the validation set after an epoch. We use batch sizes of 20 for DATA-S and 100 for DATA-L. We train for 50 epochs on DATA-S and for 25 epochs on DATA-L, picking the best-performing model on the validation set. Parameters of the models are randomly initialized uniformly in $[-0.05, 0.05]$, except the forget bias of the word-level LSTM, which is initialized to $1$. For regularization we use dropout \cite{hinton2012improving} with probability 0.5 between word-level LSTM layers and on the hidden-to-output softmax layer. We clip the norm of the gradients (normalized by minibatch size) at 5. These choices were guided by previous work on word-level language modeling with LSTMs \cite{zaremba2014recurrent}.

To speed up training on DATA-L we use a sampled softmax \cite{DBLP:journals/corr/JeanCMB14} with the number of samples equal to 20\% of the vocabulary size \cite{DBLP:conf/acl/ChenGA16}. 
Although \newcite{kim2016character} used a hierarchical softmax \cite{morin2005hierarchical} for the same purpose, a recent study \cite{grave2016efficient} shows that it is outperformed by sampled softmax on the Europarl corpus, from which DATA-L was derived \cite{botha2014compositional}.

\noindent\textbf{RHN-based models} are optimized as in \newcite{zilly2016recurrent}, except that we unrolled the networks for 70 time steps in truncated BPTT, and dropout rates were chosen to be as follows: 0.2 for the embedding layer,
0.7 for the input to the gates, 0.7 for the hidden units and 0.2 for the output activations.

\section{Sizes and speeds}\label{size_speed}
\begin{figure}[h]
\centering
\includegraphics[width=0.49\columnwidth]{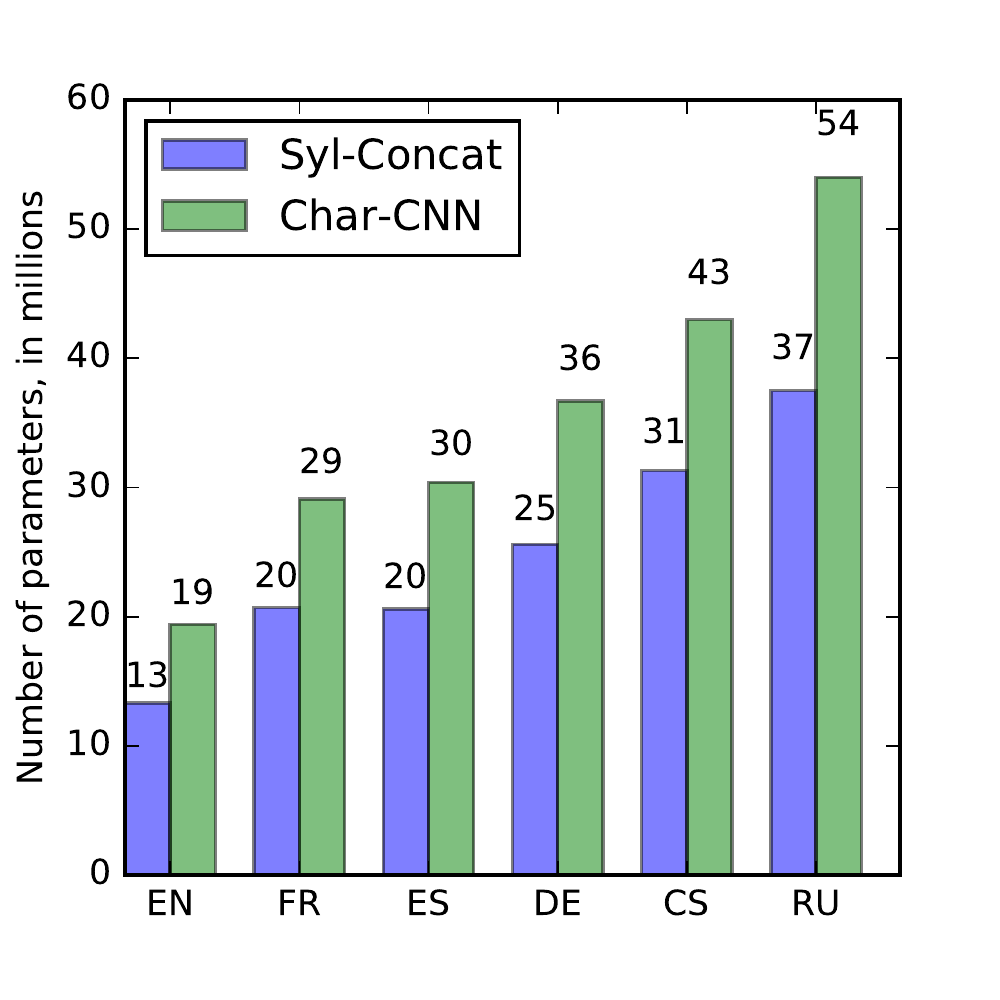}
\includegraphics[width=0.49\columnwidth]{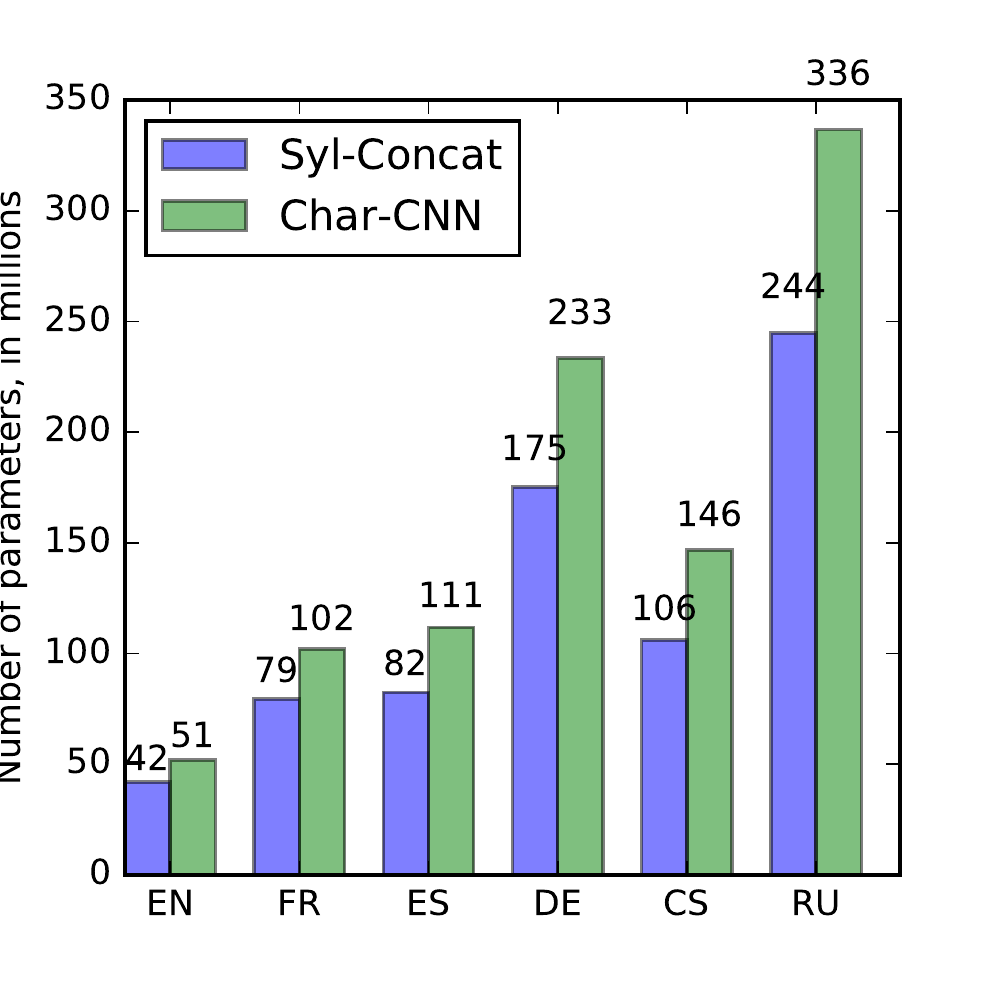}
\caption{Model sizes on DATA-S (left) and DATA-L, in millions of trainable variables.}
\label{sizes}
\end{figure}

On DATA-S, Syl-Concat has 28\%--33\% fewer parameters than Char-CNN, and on DATA-L the reduction is 18\%--27\% (see Fig. \ref{sizes}).

\noindent Training speeds are provided in the Table \ref{speed}. Models were implemented in TensorFlow, and were run on NVIDIA Titan X (Pascal). 
\begin{table}[htbp]
\setlength{\tabcolsep}{5pt}
\begin{center}
\begin{small}
\begin{tabular}{l c c c c c c c}
\hline
 Model   & EN & FR & ES & DE & CS & RU\\
\hline
 Char-CNN   & 9 & 8 & 8 & 7 & 6 & 6 & \parbox[t]{0.5mm}{\multirow{2}{*}{\rotatebox[origin=c]{90}{S}}}\\
 Syl-Concat & 14 & 12 & 12 & 11 & 10 & 9 \\
\hline
Char-CNN & 10 & 8 & 7 & 5 & 7 & 4 & \parbox[t]{0.5mm}{\multirow{2}{*}{\rotatebox[origin=c]{90}{L}}}\\
Syl-Concat & 22 & 13 & 13 & 6 & 10 & 5 & \\
\hline
\end{tabular}
\end{small}
\end{center}
\caption{Training speeds, in thousands of tokens per second.}
\label{speed}
\end{table}

\section*{Acknowledgements}
We gratefully acknowledge the NVIDIA Corporation for their donation of the Titan X Pascal GPU used for this research. The work of Bagdat Myrzakhmetov has been funded by the Committee of Science of the Ministry of Education and Science of the Republic of Kazakhstan under the targeted program O.0743 (0115PK02473). The authors would like to thank  anonymous reviewers and Aibek Makazhanov for valuable feedback, Makat Tlebaliyev and Dmitriy Polynin for IT support, and Yoon Kim for providing the preprocessed datasets.

\bibliography{lstm-syl}
\bibliographystyle{emnlp_natbib}

\end{document}